\documentclass{article}
\usepackage{spconf,amsmath,graphicx}
\usepackage{url}
\usepackage{subfigure}
\usepackage{threeparttable}
\usepackage{amsmath,bm}
\usepackage{amsfonts,amssymb} 

\title{Model Attribution of Face-swap Deepfake Videos}
\name{Shan Jia$^{\star}$ \qquad Xin Li$^{\dagger}$ \qquad Siwei Lyu$^{\star}$}
  
\address{$^{\star}$University at Buffalo, State University of New York, NY, USA \\
      $^{\dagger}$West Virginia University, WV, USA}
\begin{document}
%
\maketitle
\begin{abstract}
AI-created face-swap videos, commonly known as {\em Deepfakes}, have attracted wide attention as powerful impersonation attacks. Existing research on Deepfakes mostly focuses on binary detection to distinguish between real and fake videos. However, it is also important to determine the specific generation model for a fake video, which can help attribute it to the source for forensic investigation. In this paper, we fill this gap by studying the model attribution problem of Deepfake videos. We first introduce a new dataset with DeepFakes from Different Models (DFDM) based on several Autoencoder models. Specifically, five generation models with variations in encoder, decoder, intermediate layer, input resolution, and compression ratio have been used to generate a total of $6,450$ Deepfake videos based on the same input. Then we take Deepfakes model attribution as a multiclass classification task and propose a spatial and temporal attention based method to explore the differences among Deepfakes in the new dataset. Experimental evaluation shows that most existing Deepfakes detection methods failed in Deepfakes model attribution, while the proposed method achieved over 70\% accuracy on the high-quality DFDM dataset\footnote{The DFDM dataset and codes are available from \url{https://github.com/shanface33/Deepfake_Model_Attribution}}. 
\end{abstract}
\vspace{-0.25cm}
\section{Introduction}
\label{sec:intro}
\vspace{-0.25cm}
The term {\em Deepfake}, a portmanteau of `deep learning’ and `fake’, is often used to refer to the AI-synthesized face-swap videos, in which the faces of the original subject (the `source') are replaced with those of the other person (the `target') generated using deep learning models with the same facial expressions of the source. Since the nascence of the first Deepfakes in late 2017, they have become increasingly easier to produce thanks to the availability of open-source generation tools. Deepfakes with manipulated identities have raised wide concerns \cite{lyu2020deepfake}, especially when they are weaponized by malicious actors to target individuals or spread disinformation \cite{chesney2019deep}.

\begin{figure}[t]
\begin{center}
\includegraphics[width=3.23in]{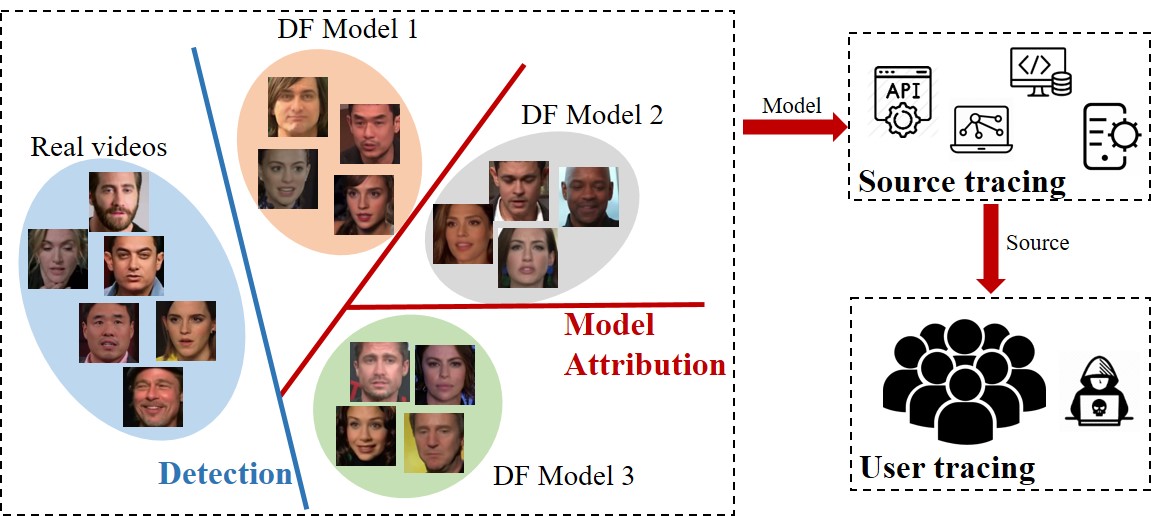}
\vspace{-0.25cm}
\caption{
\small Different from existing detection methods to distinguish between real and fake videos, Deepfake model attribution aims to identify the generation model of a particular DeepFake video, which may be used to further trace its author and source.}
\vspace{-0.75cm}
\label{fig:1}  
\end{center}
\end{figure}
Accordingly, we have seen a growing research interest in Deepfake detection methods in recent years. The DeepFake detection methods exploit visual/signal artifacts~\cite{li2019exposing, yang2019exposing, mittal2020emotions,guarnera2020deepfake} and/or employ novel deep neural networks \cite{afchar2018mesonet, nguyen2019capsule, de2020deepfake, kim2021fretal}. Correspondingly, many datasets of Deepfake videos have been created in the past few years, notable examples including
Faceforensics++~\cite{rossler2019faceforensics}, Celeb-DF~\cite{li2020celeb}, DFDC~\cite{dolhansky2019deepfake}, Deeperforensics-1.0~\cite{jiang2020deeperforensics}, and WildDeepfake~\cite{zi2020wilddeepfake}. 

Although the state-of-the-art Deepfake detection methods have demonstrated encouraging performance on benchmark datasets, the specific means by which the DeepFakes were created are also important for forensic analysis (see Fig. \ref{fig:1}). For instance, knowing that the synthesis model is a copy of a publicly available tool can narrow down the list of users who have downloaded it. 
To this end, we need a general and flexible method for Deepfake generation model attribution to determine which synthesis model was used to create the Deepfake video. 

The model attribution problem has been considered in recent studies~\cite{yu2019attributing, goebel2020detection, girish2021towards} in the case of Generative Adversarial Network (GAN)-based models. However, the model attribution method designed for GAN images cannot be extended to  face-swap Deepfakes videos. The latter is more challenging because the Autoencoder based models to create most Deepfakes~\cite{dolhansky2020deepfake} will attenuate high-frequency features in the generated video frames, on which the previous methods for GAN model attribution predicate. 

In this work, we provide a new method for the Deepfakes model attribution. As the majority of face-swap videos are created with tools (i.e., \cite{fs, dfl}] based on the autoencoder models~\cite{dolhansky2020deepfake}, which have higher quality in video than swaps generated with GAN-based models and require much less fine-tuning~\cite{dolhansky2020deepfake}, we focus on such models in this work. We address two questions: 1) do different generation models result in visually similar but statistically distinguishable Deepfake videos? and 2) since several open-source generation models have been widely used to create Deepfake videos, are the differences among variations of the same generation model consistent and detectable from the input video? 

\begin{figure}[t]
\begin{center}
\includegraphics[width=3.3in]{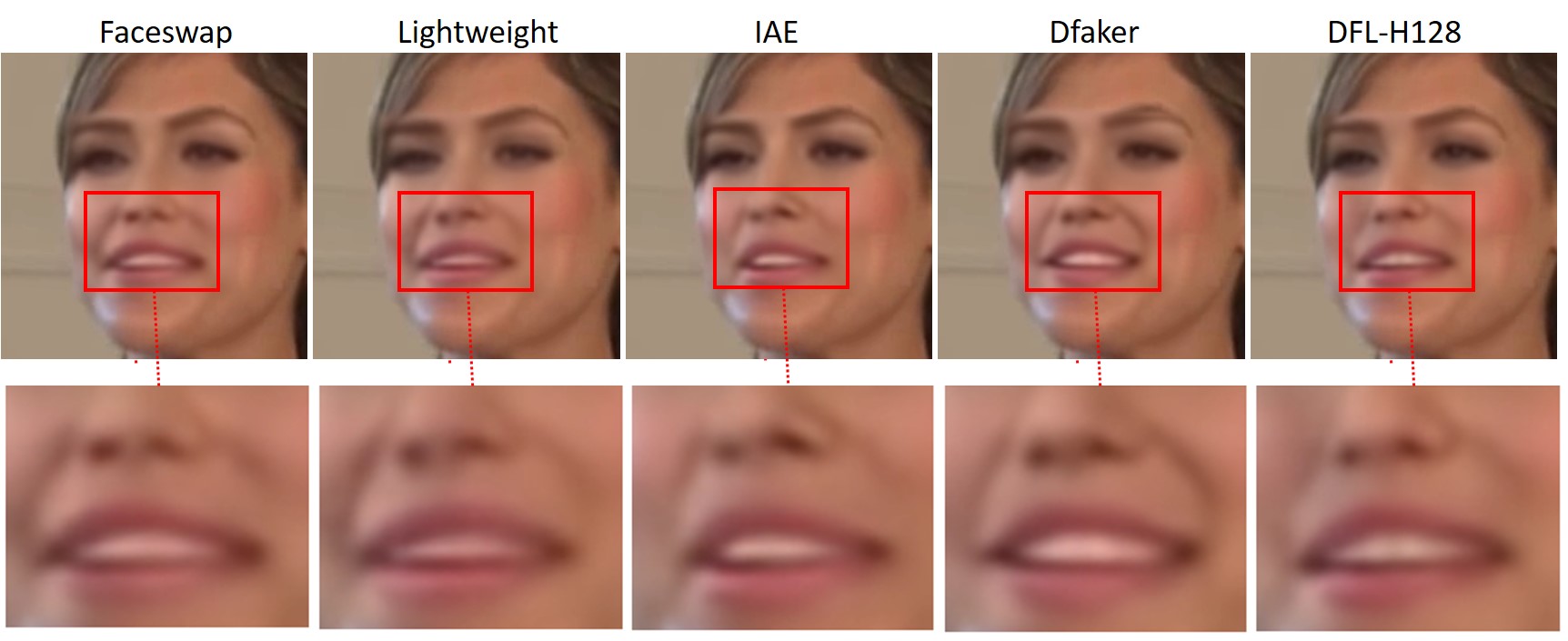}
\vspace{-0.35cm}
\caption{\small Examples of Deepfakes generated by different models in our DFDM dataset. Visible but subtle visual differences can be observed in face regions. Best view when zoom in.}
\vspace{-0.65cm}
\label{fig:2}  
\end{center}
\end{figure}
To answer these questions, we first construct a new dataset as {\em \underline{D}eep\underline{F}akes generated from \underline{D}ifferent \underline{M}odels} (DFDM). Specifically, based on the most popular off-the-shelf software, \textit{Facewap}~\cite{fs},
we have created five categories of balanced Deepfake videos using Autoencoder models with variations in the encoder, decoder, intermediate layer, and input resolution. 
For the second question, we formulate the face-swap Deepfakes model attribution as a multi-classification problem and design a novel Deepfakes model attribution method by extracting subtle and discriminative features based on spatial and temporal attention.

The main contributions of our work can be summarized as follows. 1) We aim to tackle the problem of fine-grained model attribution for face-swap Deepfake videos. To the best of our knowledge, this is the first method to solve this problem. 2) We provide a new dataset DFDM that highlights subtle visual differences caused by different Deepfakes generation models and indistinguishable to human eyes (see Fig.\ref{fig:2}). 3) A simple and effective Deepfakes model attribution method based on {\em spatial and temporal attention}, named DMA-STA, is designed and evaluated on the DFDM dataset, achieving over 70\% accuracy in identifying the high-quality Deepfakes. 

\vspace{-0.25cm}
\section{Related work}     \label{sec:rel}
\vspace{-0.2cm}
\textbf{Model attribution.}
Model attribution aims to identify the specific model behind a synthetic image or video~\cite{girish2021towards}. With an increasing number of GAN models introduced in image manipulation, several studies~\cite{yu2019attributing, marra2019gans, goebel2020detection, girish2021towards, asnani2021reverse} have investigated GAN model attribution in fake images. These methods propose to extract unique artificial fingerprints left by different GAN models and perform multi-class classification for GAN model attribution. However, as far as we know, there is no previous work on Deepfakes model attribution method that works for face-swap videos. Although GAN models can be used for Deepfakes generation, they seem to work well in limited settings with even lighting conditions~\cite{dolhansky2020deepfake}. 
Consequently, most publicly available face-swap Deepfakes were created with Deepfake Autoencoder (DFAE) methods. Will variant DFAE models differentiate Deepfakes attribution? Will image-based GAN model attribution also work for DFAE with videos? Answering these questions is valuable for research on Deepfake model attribution.

\begin{figure*}[th]
\begin{center}
\includegraphics[width=6.34in]{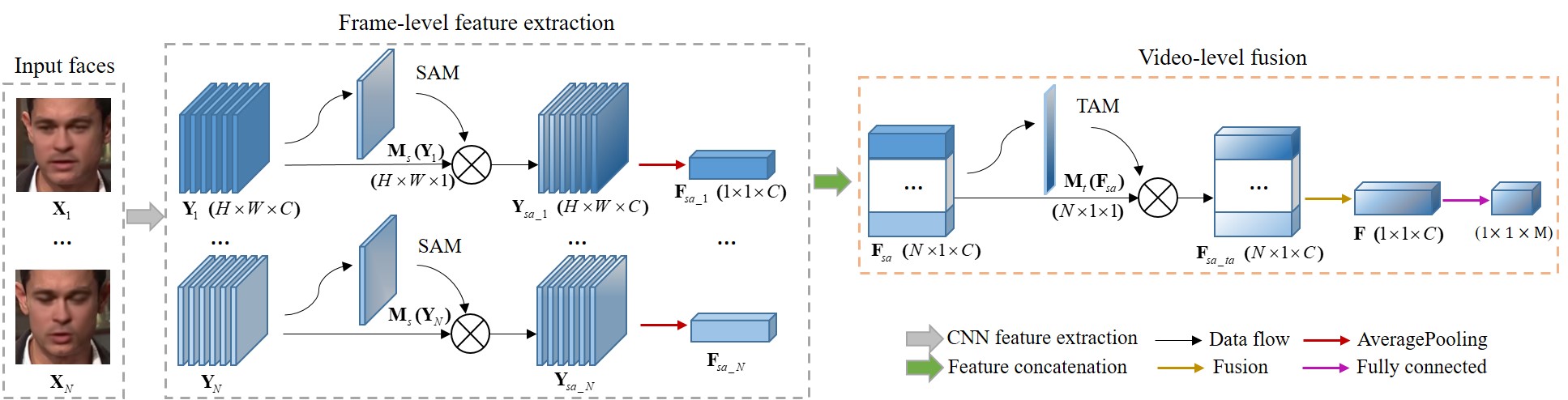}
\vspace{-0.3cm}
\caption{\small Framework of the proposed DMA-STA method. Two major components are included: 1) Feature extraction from multiple single frames based on SAM (Spatial Attention Map); 2) Video-level fusion module based on TAM (Temporal Attention Map).}
\vspace{-0.85cm}
\label{fig:4}  
\end{center}
\end{figure*}

\noindent \textbf{Deepfakes detection.}
Research on Deepfake detection involves both data generation and method design. There have been several publicly available datasets with both real video and Deepfakes proposed for binary detection. Most of them created {\em face-swap} Deepfakes with one manipulation model, or provided imbalanced Deepfakes without model annotations (as shown in Table \ref{tab:1}). 
Based on these datasets, many Deepfake detection methods have been developed. One popular category explores various artifacts, including head poses~\cite{yang2019exposing}, emotion perception~\cite{mittal2020emotions} etc., and signal-level artifacts in visual cues~\cite{li2019exposing, mittal2020emotions,guarnera2020deepfake}, and frequency domain~\cite{durall2019unmasking}, etc. Another category employs deep neural networks, such as Capsule network~\cite{nguyen2019capsule}, ensemble of CNNs~\cite{bonettini2021video}, and attention schemes~\cite{zhao2021multi}. For Deepfake model attribution, artifacts-based detection methods may fail due to the fact that the artifacts to distinguish Deepfakes from real videos exist in all Deepfakes from different models. In this paper, we will design a deep neural network-based method to learn subtle differences in Deepfake model attribution, and also evaluate how Deepfake detection methods will perform in model attribution task.
\vspace{-0.25cm}
\begin{table}[tbp]
  \centering
  \footnotesize
  \caption{Comparison of various Deepfake datasets}
  \setlength{\tabcolsep}{1.1mm}{
    \begin{tabular}{ccccc}
    \hline
    \textbf{Dataset} & \textbf{\#Deepfakes} & \textbf{\#Sub} & \textbf{\#Models} & \textbf{Model label?} \\
    \hline
    UADFV~\cite{yang2019exposing} & 49    & 49    & 1     & - \\
    \hline
    Deepfake-TIMIT~\cite{korshunov2019vulnerability} & 640   & 43    & 1     & - \\
    \hline
    FaceForensics++~\cite{rossler2019faceforensics} & 3,000 & 977   & 1     & - \\
    \hline
    DFDC Preview~\cite{dolhansky2019deepfake} & 5,244 & 66    & 2     & No \\
    \hline
    DeepfakeDetection~\cite{Dufour2019} & 3,068 & 28    & -     & No \\
    \hline
    Celeb-DF~\cite{li2020celeb} & 5,639 & 59    & 1     & - \\
    \hline
    DeeperForensics-1.0~\cite{jiang2020deeperforensics} & 10,000 & 100   & 1     & - \\
    \hline
    DFDC~\cite{dolhansky2020deepfake}  & 104,500 & 960   & 6     & No \\
    \hline
    WildDeepfake~\cite{zi2020wilddeepfake} & 3,509 & -     & -     & No \\
    \hline
    \textbf{DFDM (ours)} & \textbf{6,450} & \textbf{59}    & \textbf{5}     & \textbf{Yes} \\
    \hline
    \end{tabular}}%
    \vspace{-0.45cm}
  \label{tab:1}%
\end{table}%

\vspace{-0.15cm}
\section{Method}   \label{sec:met}
\vspace{-0.25cm}
\subsection{The DFDM dataset}
\vspace{-0.28cm}
As there are no existing datasets with labeled Deepfakes generated from various models, we create a new DFDM dataset for Deepfakes model attribution. Considering that most public Deepfakes were produced based on Autoencoder architectures~\cite{dolhansky2020deepfake, zi2020wilddeepfake}, we focus on Autoencoder models for high-quality Deepfakes generation. 

We use the most popular open-source software \textit{Facewap} \cite{fs} with several optional DFAE models, including the original model (Faceswap) created by the reddit user, and models from \textit{DeepFaceLab}~\cite{dfl}. As the first attempt in Deepfake video model attribution, we elaborately selected five models based on the following criteria: use the original Faceswap model as the baseline to select models with only one variation (in encoder, decoder, intermediate layer, and input resolution, respectively) for the most subtle model attribution. Table \ref{tab:2} shows the details of the five models, including Faceswap~\cite{fs}, Lightweight~\cite{fs}, IAE~\cite{fs}, Dfaker~\cite{dfaker}, and DFL-H128~\cite{perov2020deepfacelab}. Note that other models provided by \textit{Facewap} have over one variation with each other, which we believe are easier to identify than the five models we selected here.  

\begin{table}[tbp]
\newcommand{\tabincell}[2]{\begin{tabular}{@{}#1@{}}#2\end{tabular}} %
  \centering
  \footnotesize
  \vspace{-0.2cm}
  \caption{Structures of Deepfakes generation models used in the proposed DFDM dataset}
    \setlength{\tabcolsep}{0.6mm}{
  \begin{threeparttable}
    \begin{tabular}{cccccc}
    \hline
    \textbf{Model} & \textbf{Input} &\textbf {Output} & \textbf{Encoder} & \textbf {Decoder} & \textbf {Variation} \\
    \hline
    \tabincell{c}{Faceswap \\(baseline)} & 64    & 64    & 4Conv+1Ups & 3Ups+1Conv & / \\
    \hline
    Lightweight & 64    & 64    & 3Conv+1Ups & 3Ups+1Conv & Encoder \\
    \hline
    IAE  & 64    & 64    & 4Conv & 4Ups+1Conv & \tabincell{c}{Intermediate\\layers; Shared \\Encoder\&Decoder} \\
    \hline
    Dfaker & 64    & 128   & 4Conv+1Ups & \tabincell{c}{4Ups+3Resi-\\dual+1Conv} & Decoder \\
    \hline
    DFL-H128  & 128   & 128   & 4Conv+1Ups & 3Ups+1Conv & Input resolution \\
    \hline
    \end{tabular}%
    \begin{tablenotes}   
        \footnotesize        
    \vspace{-0.05cm}
        \item[] `Conv' - convolution layer, and `Ups' - upscaling function.
      \end{tablenotes}  
    \end{threeparttable}}
     \vspace{-0.45cm}
  \label{tab:2}%
\end{table}%


To generate videos with high diversity, we chose the real videos in Celeb-DF dataset~\cite{li2020celeb}, which contains 590 YouTube interviews of 59 celebrities. 
We used the S3FD detector~\cite{zhang2017s3fd} and FAN face aligner~\cite{bulat2017far} for face extraction, trained each model with 100,000 iterations, and generated the final Deepfakes after face convert in MPEG4.0 format. Three H.264 compression rates are considered to get videos with different qualities, including lossless with the constant rate factor (crf) as 0, high quality with crf as 10, and low quality with crf as 23. Totally, 6,450 Deepfakes have been created in the DFDM dataset. Fig. \ref{fig:2} shows the face examples in Deepfakes from five models based on the same training data. The visual differences in face regions demonstrate evidence of model attribution artifacts.

\vspace{-0.2cm}
\subsection{The DMA-STA method}
\vspace{-0.2cm}
To explore if the observed artifacts among different Deepfakes are consistent and detectable, we design a model attribution method (named DMA-STA) by fusing spatial (frame-level) and temporal (video-level) attention schemes for discriminative feature extraction. 

\noindent \textbf{Frame-level feature extraction.} 
We first extract frame-level features from \textit{N} cropped faces $\left\{\textbf{X}_i, i\in [1,N]\right\}$. As shown in Fig.\ref{fig:4}, a CNN model is employed to automatically extract high-level representations $\left\{\textbf{Y}_i \in \mathbb{R}^{H\times W \times C}\right\}$ from each face ${\textbf{X}_i}$. Next, the 2D Spatial Attention Map (SAM) $\left\{\textbf{M}_s(\textbf {Y}_i) \in \mathbb{R}^{H\times W \times 1}\right\}$ is obtained using the lightweight and flexible convolutional block attention module proposed by \cite{woo2018cbam}. Specifically, the SAM is computed as:
\vspace{-0.2cm}
\begin{equation}
\textbf{M}_s(\textbf {Y}_i) = \sigma(f^{7\times 7}([AvgPool(\textbf{Y}_i)];[MaxPool(\textbf{Y}_i)]))
\end{equation}

\vspace{-0.15cm}
\noindent where $\sigma$ denotes the sigmoid function to obtain probabilities to weigh the feature maps, $f^{7\times7}$ is a convolution operation with the filter size of $7\times7$, $AvgPool$ and $MaxPool$ represent the average pooling and max pooling respectively to highlight the information regions, and `;' means feature concatenation. Based on the SAM, the adaptive features $\left\{\textbf{Y}_{sa\_i} \in \mathbb{R}^{H\times W \times C}\right\}$ are obtained by,
\vspace{-0.2cm}
\begin{equation}
\textbf{Y}_{sa\_i} = \textbf{M}_s(\textbf {Y}_i)\otimes {\mathbf{Y}_i}
\vspace{-0.2cm}
\end{equation}
\noindent where $\otimes$ represents the element-wise multiplication. Note that the SAM can be integrated into each subblock in CNN, such as the ResBlock in ResNet. To aggregate the frame features, extracted features $\textbf{Y}_{sa\_i}$ are finally fed into the average pooling layer, outputting the final frame features  $\left\{\textbf{F}_{sa\_i} \in \mathbb{R}^{1\times 1 \times C}\right\}$.

\noindent \textbf{Video-level fusion.} Previous studies on video-level Deepfakes detection mostly fused multi-frame features based on averaging the network predictions (score fusion)~\cite{li2019exposing, afchar2018mesonet, nguyen2019capsule, korshunov2019vulnerability, li2021frequency}. To improve the classification performance, we introduce the temporal attention map (TAM) to adaptively aggregate the features from different frames. The frame features $\textbf{F}_{sa\_i}$ from \textit{N} faces are first concatenated to get the multi-frame representations $\left\{\textbf{F}_{sa} \in \mathbb{R}^{N\times 1 \times C}\right\}$, which are fused to get the adaptive representation $\left\{\textbf{F}_{sa\_ta} \in \mathbb{R}^{N\times 1 \times C}\right\}$, computed as,
\vspace{-0.2cm}
\begin{equation}
\textbf{F}_{sa\_ta} = \textbf{M}_t(\textbf{F}_{sa})\otimes {\mathbf{F}_{sa}}
\vspace{-0.1cm}
\end{equation}
\noindent where $\left\{\textbf{M}_t(\textbf{F}_{sa}) \in \mathbb{R}^{N\times1 \times 1}\right\}$ is the temporal attention map, using the similar structure to SENet~\cite{hu2018squeeze},

\vspace{-0.25cm}
\begin{equation}
\textbf{M}_t(\textbf{F}_{sa}) = \sigma f_2(\delta f_1[AvgPool(\textbf{F}_{sa})])
\vspace{-0.1cm}
\end{equation}
\noindent where $\delta$ is the ReLU function, and $f_1$ and $f_2$ are two fully connected layers. Let the elements in the $j$ channel of $\textbf{F}_{sa\_ta}$ and $\textbf{M}_t(\textbf{F}_{sa})$ as $f_{i,j}$ and ${m}_{i}$, respectively; then the final representation $\left\{\textbf{F} \in \mathbb{R}^{1\times 1 \times C}\right\}$ is formulated as Equation (5), which is finally fed into the fully connected layer for classification (outputting the class probabilities for $M$ category of Deepfakes).
\vspace{-0.2cm}
\begin{equation}
\textbf{F}_j = \frac{\sum_{i=1}^{N} f_{i,j}} {\sum_{i=1}^{N} {m}_{i}} \;\;\; \;(j \in  [1,C])
\vspace{-0.1cm}
\end{equation}

The cross-entropy loss is used to train the network. 
Different from existing Deepfake detection methods, we design the attention mechanism in both frame and video level for subtle difference extraction in model attribution. When compared to previous fingerprint based GAN model attribution studies, the DMA-STA method directly takes the whole faces as input instead of designing auxiliary artifacts for the multi-classification based task, so that the differences among different Deepfakes can be learned effectively and automatically.

\vspace{-0.25cm}
\section{Experiments}   \label{sec:exp}
\vspace{-0.2cm}
\subsection{Experimental Settings}
\vspace{-0.2cm}
We use the ResNet-50~\cite{he2016deep} as the CNN feature extractor in DMA-STA.
The videos in DFDM are randomly split into a training set (70\%) and a testing set (30\%). To balance the distribution of faces in frame selection, we selected 10 frames of each video by periodic sampling for the proposed video-level attribution scheme. Consequently, the network takes the cropped faces in $224\times224\times3\times\ 10$ as input. To train the network, the optimizer is set to SDG with the weight decay of $5\times 10^{-4}$, and momentum equal to 0.9. The initial learning rate of 0.01 is divided by 10 every 40 epochs to 0.0001. A batch size of 10 is used for training with 300 epochs. 

Several existing classification methods with open-source codes are re-trained and evaluated on the DFDM dataset, including CNN based detection methods~\cite{afchar2018mesonet, nguyen2019capsule, rossler2019faceforensics, de2020deepfake}, artifacts based detection~\cite{li2019exposing, durall2019unmasking}; and GAN image model attribution methods~\cite{asnani2021reverse, marra2019gans}. 
Classification accuracy is used as the evaluation metric. Note that for a fair comparison, we implemented all these methods in video-level fusion using the same frame number (10) as the proposed DMA-STA. 

\vspace{-0.35cm}
\subsection{Comparison with existing methods}
\vspace{-0.2cm}
\textbf{Attention schemes}. Table \ref{tab:3} compares our method with different attention schemes on the high-quality videos in DFDM. Our proposed DMA-STA achieves the best overall accuracy of 71.94\%. Further, we can observe that these methods show better performance in identifying Deepfakes generated by models with decoder variations, i.e., Dfaker, and DFL-H128.

\vspace{-0.5cm}
\begin{table}[hbp]
\newcommand{\tabincell}[2]{\begin{tabular}{@{}#1@{}}#2\end{tabular}} %
  \centering
  \footnotesize
  \caption{Comparison of different attention schemes (\%)}
  \setlength{\tabcolsep}{1.1mm}{
    \begin{tabular}{c|ccccc|c}
    \hline
    \multicolumn{1}{c|}{\textbf{Attention}} & \multicolumn{1}{c}{FS} & \multicolumn{1}{c}{LW} & \multicolumn{1}{c}{IAE} & \multicolumn{1}{c}{Dfaker} & \multicolumn{1}{c|}{DFL} & \multicolumn{1}{c}{Overall}  \\
    \hline
    ResNet-50~\cite{he2016deep} & 54.84 & 57.36 & 70.54 & {89.92} & 70.54 & 68.02  \\
    \hline
    CBAM~\cite{woo2018cbam}  & 52.42 & {63.57} & 69.77 & 84.50  & 74.42 & 68.53  \\
    \hline
    SAM+FA~\cite{meng2019frame} & {64.34} & 42.64 & 76.61 & 74.42 & 79.07 & 66.82\\
    \hline
    SAM+Ave & 58.87 & 51.16 & {76.74} & 80.62 & 79.07 & 68.84 \\
    \hline
   \textbf{Ours: DMA-STA}  & 63.57 & 58.91 & 66.67 & 82.95 & {87.60}  & \textbf{71.94}  \\
    \hline
    \end{tabular}
    \begin{tablenotes}   
        \scriptsize        
    \vspace{-0.05cm}
        \item[]FS is shorted for the Faceswap model, LW is the Lightweight model, and DFL is the DFL-H128 model.
      \end{tablenotes} }
\vspace{-0.30cm}
  \label{tab:3}%
\end{table}%

\begin{table}[t]
  \centering
  \footnotesize
  \caption{Comparison of different methods on DFDM (\%)}
  \setlength{\tabcolsep}{1.2mm}{
    \begin{tabular}{c|ccccc|c}
    \hline
    \multicolumn{1}{c|}{\textbf{Method}} & \textbf{FS} & \textbf{LW} & \textbf{IAE} & \textbf{Dfaker} & \textbf{DFL} & \textbf{Overall} \\
    \hline
    MesoInception~\cite{afchar2018mesonet} & 6.98  & 2.33  & 79.07 & 79.07 & 4.65  & 20.93 \\
    \hline
  Xception~\cite{rossler2019faceforensics} & 0.77  & 0     & 12.40  & 12.40  & 19.38 & 20.93 \\
    \hline
  R3D~\cite{de2020deepfake} & 27.13 & 25.58 & 15.5  & 20.16 & 18.61 & 21.40 \\
    \hline
  DSP-FWA~\cite{li2019exposing}& 17.05 & 7.75  & 43.41 & 40.31 & 8.87  & 23.41 \\
    \hline
   DFT-spectrum~\cite{durall2019unmasking} & 99.92&	3.26&	0.23&	27.21&	48.91&	35.91
 \\
    \hline
 DnCNN~\cite{asnani2021reverse}&2.33&	0&	0&	7.75&	 99.22&	21.86
 \\
    \hline
     GAN\_Fingerprint~\cite{marra2019gans}&  20.16 &  22.48& 54.26& 21.71 & 26.36& 28.99\\
    \hline
 Capsule\cite{nguyen2019capsule} & 32.56 & 42.64 & 69.77 & 73.64 & 58.91 & 55.50 \\
    \hline
  {Ours: DMA-STA}  & 63.57 & 58.91 & 66.67 & 82.95 & 87.60  & \textbf{71.94} \\
    \hline
    \end{tabular}} 
\vspace{-0.4cm}
  \label{tab:4}%
\end{table}%

\noindent \textbf{Classification methods}. We conducted comparison experiments (all based on 10-frame feature fusion) to show how Deepfake detection and GAN model attribution methods perform in identifying Deepfakes on the high-quality DFDM dataset. Table \ref{tab:4} shows that most methods fail in identifying Deepfakes, with an overall accuracy less than 25\%, indicating the weak or inconsistent artifacts/noise patterns among different Deepfakes. Benefiting from the rich representation of the residual network and adaptive feature learning of attention mechanisms, our method achieves an overall accuracy of 71.94\%, 16\% higher than the Capsule 
network.

\vspace{-0.35cm}
\subsection{Comparison on different video qualities}
\vspace{-0.25cm}
Table \ref{tab:6} compares the model attribution performance of our method under both intra- and cross-quality testing scenarios on the DFDM. The video compression shows a significant influence on the attribution results, with the overall accuracy ranging from 23.57\% to 73.64\%. Furthermore, it can be observed that training on videos with lower qualities and testing on videos with higher quality results in better performance than vice versa, e.g., HQ$\rightarrow$NoL with 70.85\% overall accuracy, comparable to HQ$\rightarrow$HQ results with 71.94\% (same trends in LQ$\rightarrow$NoL and LQ $\rightarrow$ HQ cases). 
 \vspace{-0.35cm}
\begin{table}[tbp]
  \centering
  \footnotesize
  \caption{Comparison on different video qualities (\%)}
    \begin{threeparttable}
    \begin{tabular}{c|ccccc|c}
    \hline
    \multicolumn{1}{c|}{\textbf{Train$\rightarrow$Test}}& \multicolumn{1}{c}{\textbf{FS}} & \multicolumn{1}{c}{\textbf{LW}} & \multicolumn{1}{c}{\textbf{IAE}} & \multicolumn{1}{c}{\textbf{Dfaker}} & \multicolumn{1}{c|}{\textbf{DFL}} & \multicolumn{1}{c}{\textbf{Overall}} \\
    \hline
    {NoL$\rightarrow$NoL} & 52.42 &54.26 & 82.17 & 94.6 & 86.82 & \textbf{73.64}  \\ 
    \hline
    {NoL$\rightarrow$HQ}
       & 41.09 & 42.64 & 75.00 & 61.24 & 83.72 &\textbf{60.62} \\
    {NoL$\rightarrow$LQ} 
       & 34.88 & 41.09 &  33.59 & 4.65  & 3.88  & \textbf{23.57} \\
    \hline
    {HQ$\rightarrow$HQ}
    & 63.57 & 58.91 & 66.67 & 82.95 & 87.60  & \textbf{71.94} \\
    \hline
    {HQ$\rightarrow$NoL} 
      & 58.14 & 45.74 & 82.03 & 86.82 & 82.17 & \textbf{70.85} \\
    {HQ$\rightarrow$LQ} 
       & 20.31 & 27.91 & 53.49 & 56.59 & 36.43 & \textbf{38.91} \\
    \hline
    {LQ$\rightarrow$LQ} & 32.26 & 25.58 & 59.69 & 72.09 & 69.77 & \textbf{51.63}\\
    \hline
    {LQ$\rightarrow$NoL} 
       & 50.00 & 22.48 & 78.29 & 59.69 & 52.71 & \textbf{52.56} \\
    {LQ$\rightarrow$HQ}
     & 40.62 & 28.68 & 67.44 & 72.09 & 61.24 & \textbf{53.95} \\
    \hline
    \end{tabular}%
    \begin{tablenotes}  
        \scriptsize         
    \vspace{-0.05cm}
        \item[]LQ denotes the low quality Deepfakes, HQ denotes the high quality Deepfakes, and NoL denotes the Deepfakes encoded by H.264 with no losses.
      \end{tablenotes}  
    \end{threeparttable}
  \label{tab:6}%
  \vspace{-0.55cm}
\end{table}%

\section{Conclusions}
\vspace{-0.25cm}
In this work, we have made the first attempt to explore the Deepfakes model attribution of different Autoencoder structures by creating a new dataset (DFDM) with Deepfakes from five models and designing a model attribution method for Deepfakes identification (DMA-STA). Although evaluation results show that several state-of-the-art Deepfake detection or GAN image attribution methods failed in identifying Deepfake videos, the proposed method based on both spatial and temporal attention achieves over 70\% accuracy on higher-quality Deepfakes identification. Our future work includes adding more Deepfakes generation models and enhancing the robustness of the method to video compression.

\section{Acknowledgement}
\vspace{-0.3cm}
This work is supported by the US Defense Advanced Research Projects Agency (DARPA) Semantic Forensic (SemaFor) program, under Contract No. HR001120C0123, and NSF CITeR Award 20f1x5 ``Deepfake Video Fingerprinting". The views, opinions and/or findings expressed are those of the authors and should not be interpreted as representing the official views or policies of the Department of Defense or the U.S. Government. 

\bibliographystyle{IEEEbib}
\small
\bibliography{DF}

\end{document}